\documentclass{article}
\pdfoutput=1
\usepackage{spconf,amsmath,graphicx}
\usepackage{graphicx}
\usepackage{booktabs}
\usepackage{comment}
\usepackage[pagebackref=true,breaklinks=true,colorlinks,bookmarks=false]{hyperref}
\graphicspath{{../images/}}

\newcommand{\fig}[1]{Fig.~\ref{fig:#1}}
\newcommand{\tab}[1]{Table~\ref{tab:#1}}

\usepackage{graphicx}
\usepackage{amsmath}
\usepackage{amssymb}
\usepackage{booktabs}

\usepackage{xspace}

\title{Hallucination in Object Detection --- A Study in Visual Part Verification}
\name{Osman Semih Kayhan$^{\star}$ \qquad Bart Vredebregt$^{\mathsection}$
\qquad Jan C. van Gemert$^{\star \mathsection}$}
\address{$^{\star}$Computer Vision Lab, Delft University of Technology and $^{\mathsection}$Aiir Innovations
}

\begin{document}
%
\maketitle

\begin{abstract}

We show that object detectors can hallucinate and detect missing objects; potentially even accurately localized at their expected, but non-existing, position. This is particularly problematic for applications that rely on \emph{visual part verification}: detecting if an object part is present or absent. We show how popular object detectors hallucinate objects in a visual part verification task and introduce the first visual part verification dataset: DelftBikes\footnote{\url{https://github.com/oskyhn/DelftBikes}}, which has 10,000 bike photographs, with 22 densely annotated parts per image, where some parts may be missing. We explicitly annotated an extra object state label for each part to reflect if a part is missing or intact. We propose to evaluate visual part verification by relying on recall and compare popular object detectors on DelftBikes.

\end{abstract}
\section{Introduction}

\begin{figure}[t!]
	\centering
	\includegraphics[width=\linewidth]{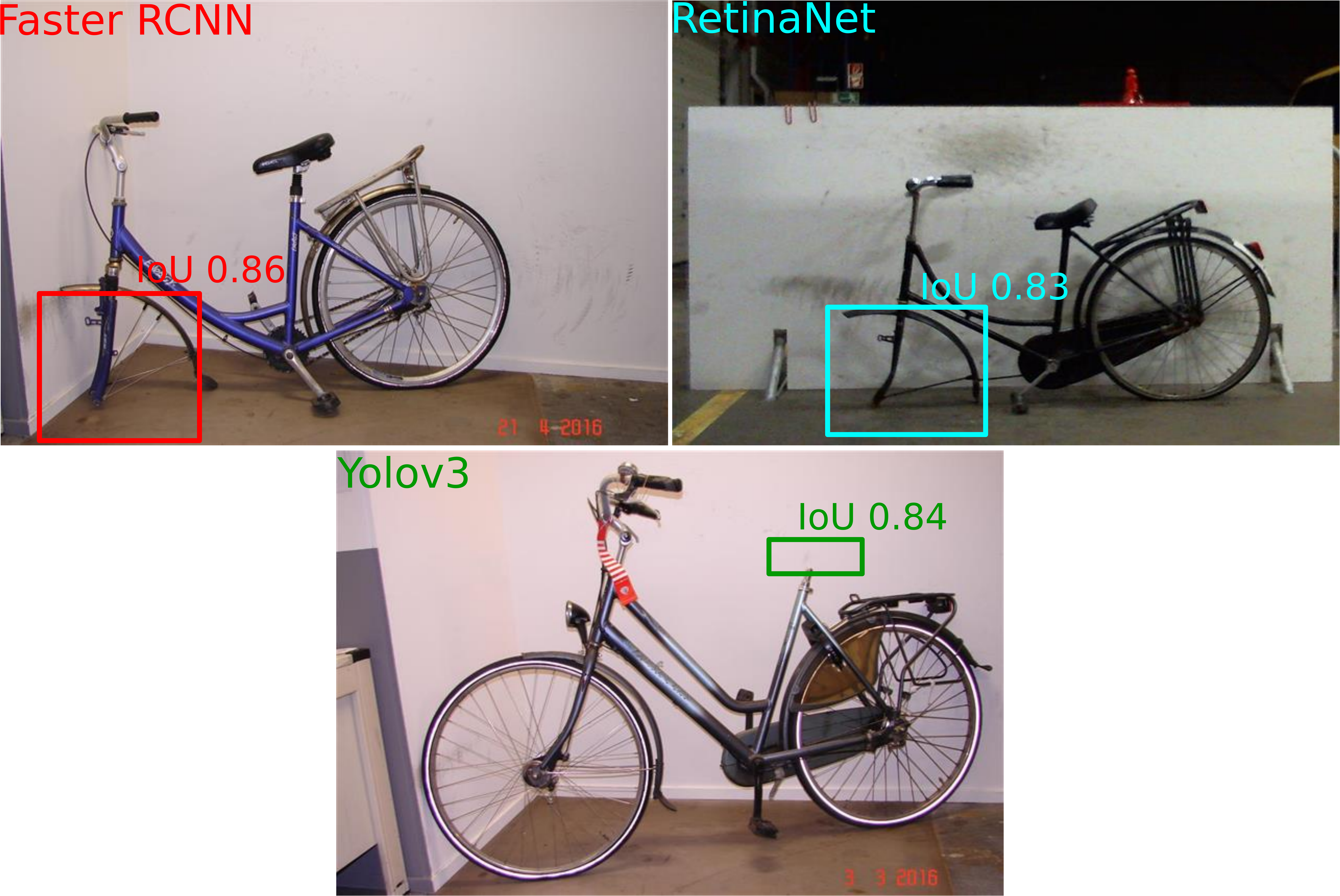}
	\caption{Hallucination examples on DelftBikes for Faster RCNN~\cite{ren2015faster}, RetinaNet~\cite{lin2017focal} and YOLOv3~\cite{redmon2018yolov3}. Faster RCNN and RetinaNet detect the front wheel and YOLOv3 predicts the saddle with a high IoU score. Deep object detectors may detect non-existent objects at their expected locations.}
	\label{fig:fig1}
\end{figure}

Automatically localizing and detecting an object in an image is one of the most important applications of computer vision. It is therefore paramount to be aware that deep object detectors can hallucinate non-existent objects, and they may even detect those missing objects at their expected location in the image, see~\fig{fig1}.  Detecting non-existing objects is particularly detrimental to applications of automatic \emph{visual part verification} or \emph{visual verification}: determining the presence or absence of an object. Examples of visual verification include infrastructure verification in map making, missing instrument detection after surgery, part inspections in machine manufacturing etc. This paper shows how popular deep detectors hallucinate objects in a case study on a novel, specifically created visual object part verification dataset: DelftBikes.

Visual verification as automatic visual inspection is typically used for manufacturing systems with applications such as checking pharmaceutical blister package~\cite{rosandich1997automated}, components on PCBs~\cite{garcia2009automated, koniar2014virtual}, solder joint~\cite{kim1999visual}, parts of railway tracks~\cite{resendiz2013automated}, rail bolts~\cite{marino2007real}, aeronautic components~\cite{ben2019automatic,san2017automatic}, objects~\cite{baerveldt2001vision}, and parts under motion~\cite{sim2006recognition}. In this paper, we do not focus on a particular application. Instead, we evaluate generic deep object detectors which potentially can be used in several visual inspection applications. 

There are important differences between visual verification and object detection. An object detector should not detect the same object multiple times.  For visual verification, however, the goal is to determine if an object is present or absent, and thus having an existing object detected multiple times is not a problem, as long as the object is detected at least once. This makes recall more important than precision. Moreover, there are differences in how much costs a mistake has. The cost for an existing object that is not detected (false negative) is that a human needs to check the detection. The cost for a missing object that is falsely hallucinated as being present (false positive) is that this object is a wrongly judged as intact and thus may cause accidents in road infrastructure, or may cause incomplete objects to be sent to a customer. The costs for hallucinating missing objects is higher than missing an existing object. These aspects motivate us to not use the evaluation measure of object detection. Object detectors are typically evaluated with mean Average Precision (mAP) and because detections of non-existent objects at lower confidence levels does not significantly impact mAP, the problem of object hallucination has largely been ignored. Here, we propose to evaluate visual verification not with precision but with a cost-weighted variant of recall.


\begin{figure*}
	\centering
	\includegraphics[width=\textwidth]{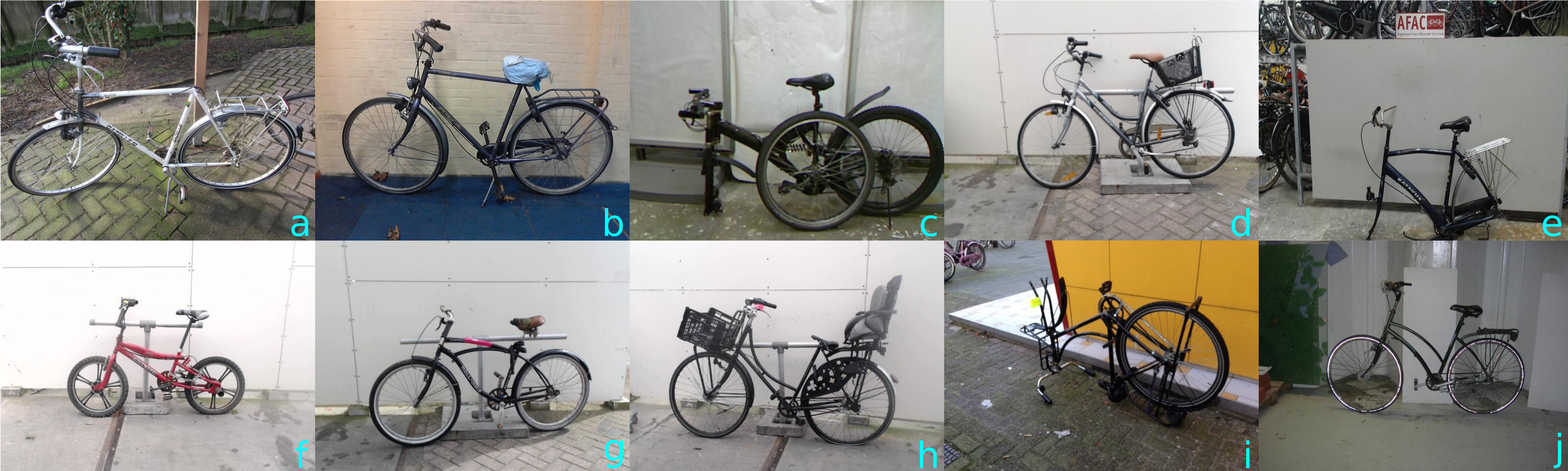}	
	\caption{Example images of our DelftBikes visual verification dataset. Each image has a single bike with 22 bounding box annotated parts. The similar pose, orientation and position can be misleading for context-sensitive detectors as often one or two parts are missing (the saddle in (a), the wheels in (e) etc.). }
	\label{fig:bike_images}
\end{figure*}

Object hallucination by deep detectors can be causes by sensitivity to the absolute position in the image~\cite{manfredi2020shift,kayhan2020translation} while also affected by scene context~\cite{Barnea_2019_CVPR, gidaris2015object,Liu_2018,zhu2017couplenet,singh2020don}. Here, we focus on the visual verification task, its evaluation measure, a novel dataset, and a comparison of popular existing detectors. Investigating context is future work.

Existing object detection datasets such as PASCAL VOC~\cite{pascal-voc-2012}, MS-COCO\cite{lin2014microsoft}, Imagenet-det~\cite{russakovsky2015imagenet}, and Open Image~\cite{DBLP:journals/corr/abs-1811-00982} have no annotated object parts. Pascal-Parts~\cite{chen2014detect} and GoCaRD~\cite{stappen2020gocard} include part labels, yet lack information if a part is missing and where, as is required to evaluate visual verification. Thus, we collected a novel visual verification dataset: DelftBikes where we explicitly annotate all part locations and part states as missing, intact, damaged, or occluded.

We have the following contributions:

1. We demonstrate hallucination in object detection for 3 popular object detectors.

2. A dataset of 10k images with 22 densely annotated parts specifically collected and labeled for visual verification. 

3. An evaluation criteria for visual verification.

\section{DelftBikes visual verification dataset}


DelftBikes (See~\fig{bike_images}) has 10,000 bike images annotated with  bounding box locations of 22 different parts where each part is in one of four possible states:\\
\textbf{intact:} The part is clearly evident and does not indicate any sign of damage. All the images in~\fig{bike_images} have an intact steer.\\
\textbf{damaged:} The part is broken or has some missing parts. In~\fig{bike_images}-g, the front part of the saddle is damaged.\\
\textbf{absent:} The part is entirely missing and is not occluded. \fig{bike_images}-e has missing front and back wheels.\\
\textbf{occluded:} The part is partially occluded because of an external object or completely invisible. The saddle in~\fig{bike_images}-b is covered with a plastic bag. 

\begin{figure}[h]
	\centering
	\includegraphics[width=\linewidth]{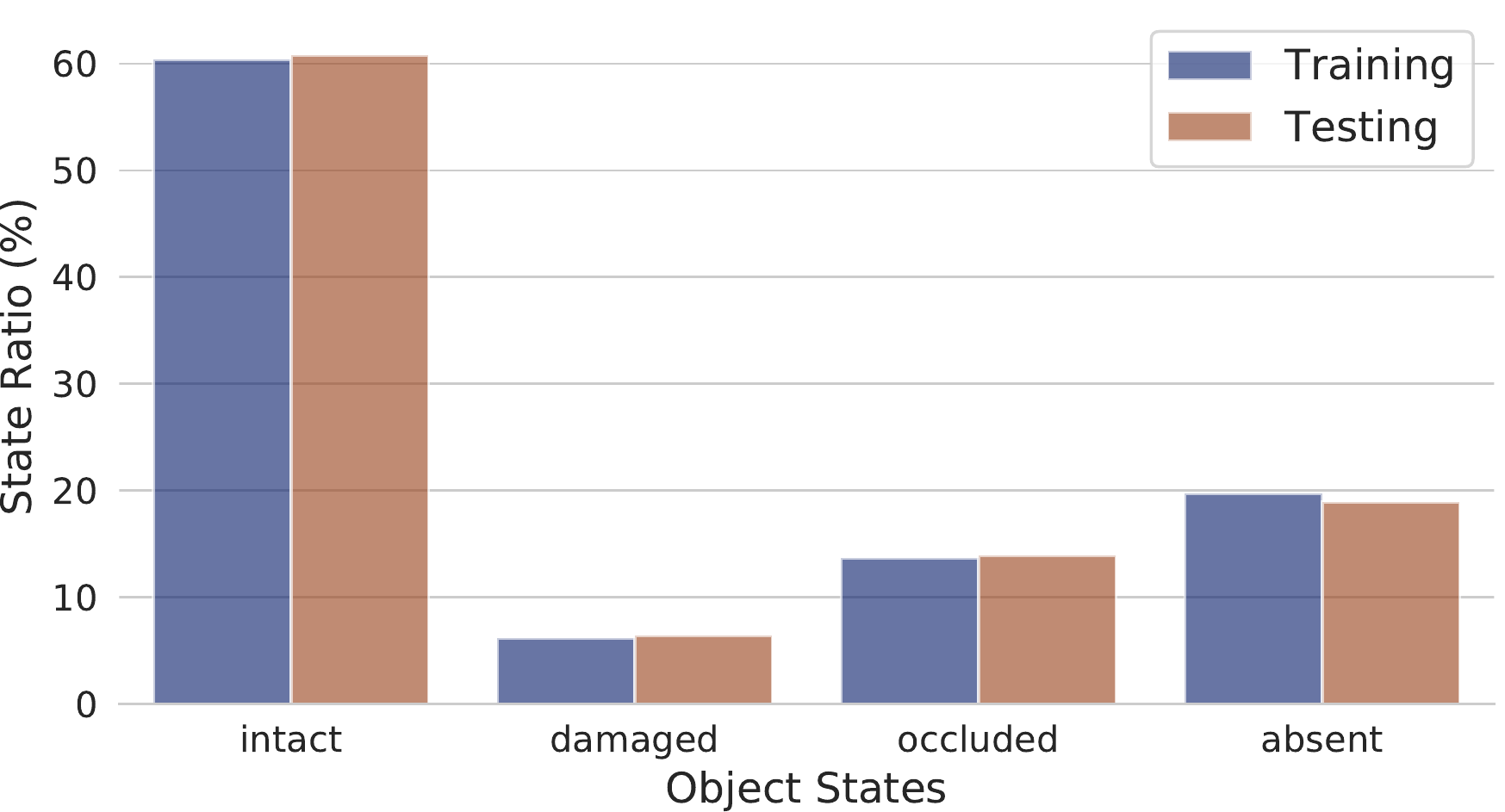}
	\caption{The distribution of part states for train and test sets in DelftBikes. The ratio of part states are roughly similar for train and test sets.
	The intact parts have the highest ratio by around $60\%$. Approximately $20\%$ of parts in the dataset are absent. The damaged and occluded parts constitute $20\%$.}
	\label{fig:states}
\end{figure}
\begin{figure}
	\centering
	\includegraphics[width=.8\linewidth]{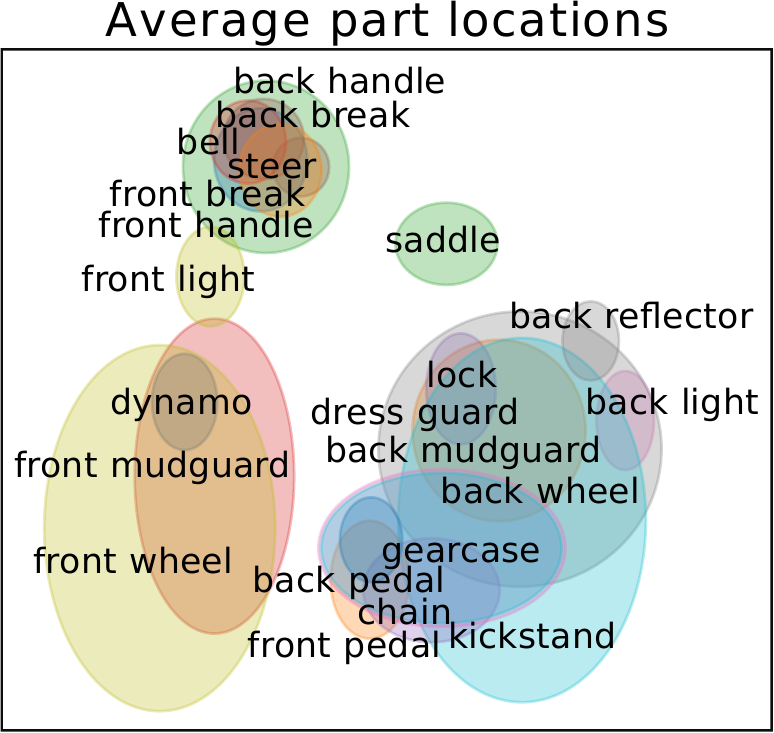}
	\caption{Averaging position and size for all 22 parts in DelftBikes resembles a bicycle, illustrating the prior in absolute position and the contextual part relations. }
	\label{fig:avg_bike}
\end{figure}

The distribution of part states is approximately similar for training and testing set, see~\fig{states}. 
The part state distribution shows 60.5\% intact, 19.5\% absent, 14\% occluded, and 6\% damaged. The \emph{front pedal}, \emph{dress guard}, \emph{chain} and \emph{back light} have respectively the highest number of intact, absent, occluded and damaged part states.
Note that even if a part is absent or occluded, we still annotate its most likely bounding box location.
DelftBikes contains positional and contextual biases. In~\fig{avg_bike} where we plot an ellipse for each part in the dataset in terms of their mean position, height and width. It is possible to recognize the shape of a bike, which indicates that there are strong part-to-part position and contextual relations. Its those biases that learning systems may falsely exploit and cause detector hallucinations.

\section{Experiments on DelftBikes}

The dataset is randomly split in 8k for training and 2k for testing.
We use a COCO pretrained models of Faster RCNN~\cite{ren2015faster} and RetinaNet~\cite{lin2017focal}. Both networks have a Resnet-50~\cite{he2016deep} backbone architecture with FPN. The networks are finetuned with DelftBikes for 10 epochs using SGD with a initial learning rate of 0.005.
The YOLOv3~\cite{redmon2018yolov3} architecture is trained from scratch for 200 epochs using SGD with an initial learning rate of 0.01. 
Other hyperparameters are set to their defaults.
We group the four part states in two categories for visual verification: (i) \textit{missing} parts consist of absent and occluded states and (ii) \textit{present} parts include intact and damaged states. 
During training, only parts with \emph{present} states are used. 

\begin{figure}[t]
	\centering
	\begin{tabular}{c@{}c}
	\includegraphics[width=0.99\linewidth]{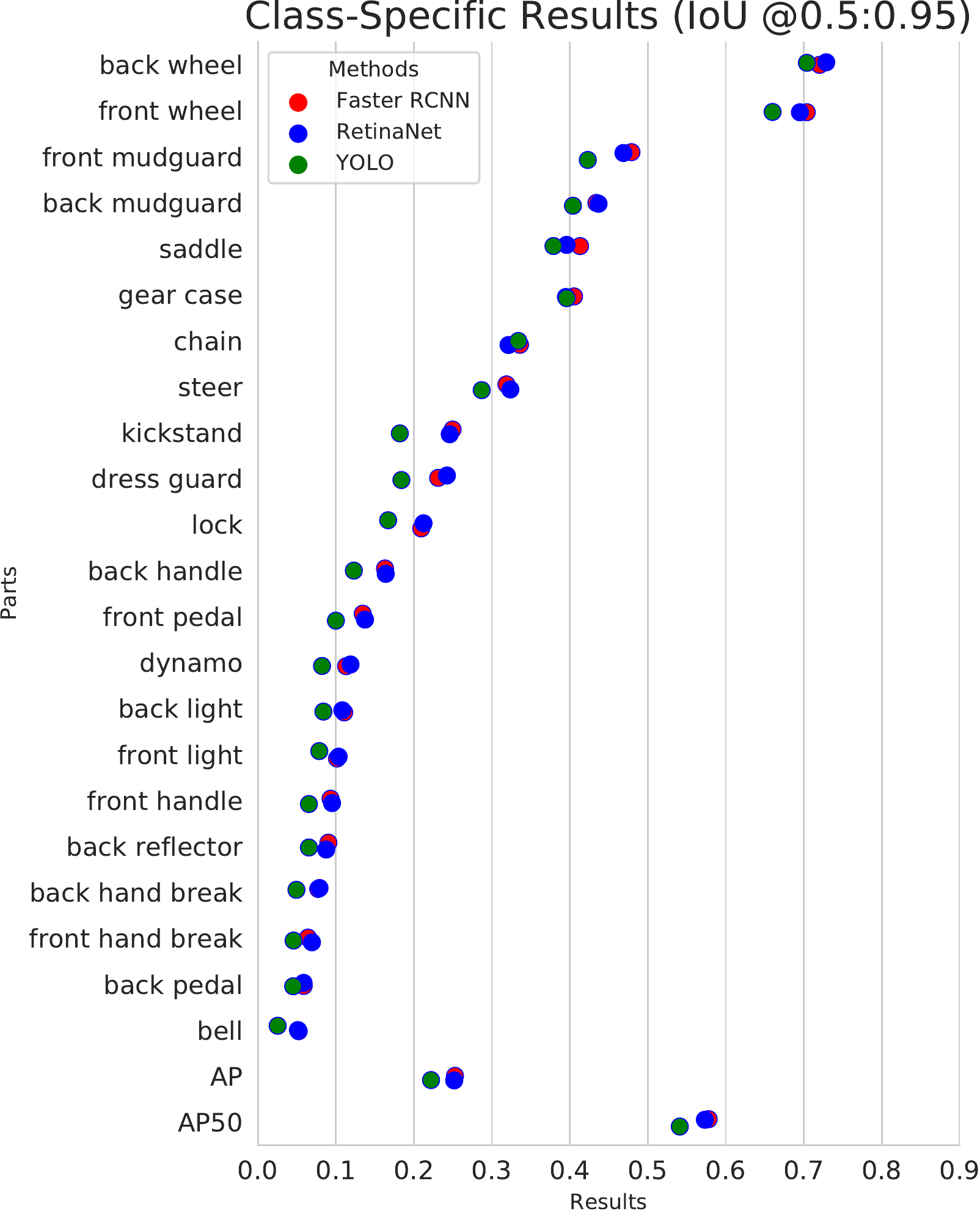}
	\end{tabular}
	\caption{Object detection results on DelftBikes. Results per category and overall performance. Notice that half of the detections are below 20\% AP score. In most of the cases, Faster RCNN and RetinaNet perform similarly and YOLOv3 is behind them.}
	\label{fig:bike_results4}
\end{figure}

\textbf{Detection.}
We first evaluate traditional object detection using AP. For object detection, the  \emph{missing} parts are not used during training nor testing. In~\fig{bike_results4}, we show results for an IoU of $0.5{:}0.95$ for the 3 detectors. For most of the classes, Faster RCNN and RetinaNet obtain approximately a similar result and YOLOv3 is a bit behind. \emph{Front wheel} and \emph{back wheel} are large and well detected. The small parts like \emph{bell} and \emph{dynamo} have under  $12\%$ AP score because they are small parts and often not present. The other parts are below $50\%$ AP, where half of the parts have less than $20\%$ AP, which makes DelftBikes already a challenging and thus interesting object detection dataset. 


\begin{figure}
	\centering
	\begin{tabular}{c@{}c}
	\includegraphics[width=.94\linewidth]
	{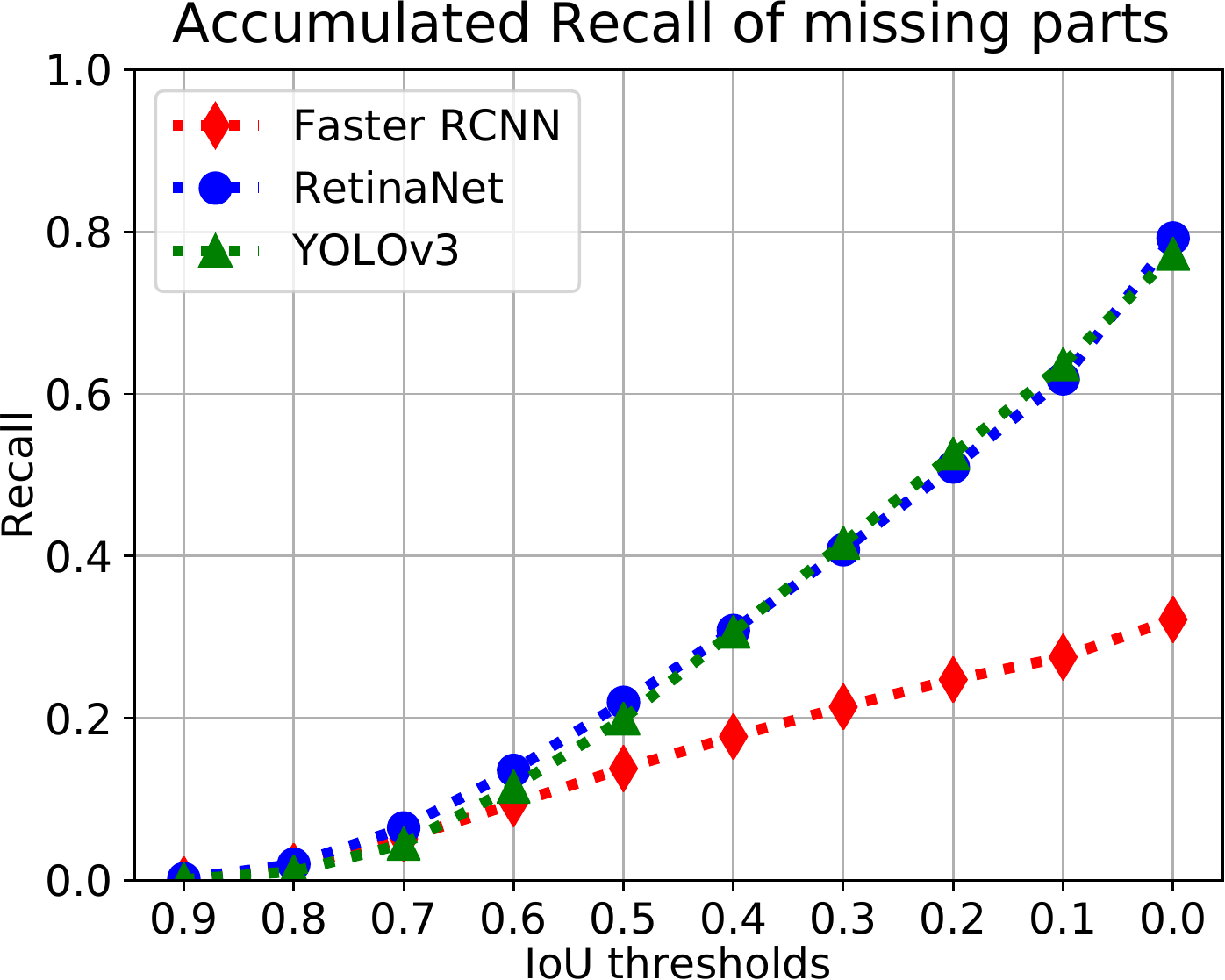}
	\end{tabular}
	\caption{Recall of \emph{missing} parts on DelftBikes for varying Intersection over Union (IoU). We annotated likely position of missing parts, and the recall of such missing parts should be as low as possible. All methods wrongly detect missing parts at approximately their expected location, as in~\fig{fig1}. }
	\label{fig:absent_recall}
\end{figure}

\textbf{Recall of \emph{missing} parts.}
Here, we analyze the hallucination failure of the detectors by evaluating how many non-existing parts they detect in an image. We calculate the IoU score for each detected \emph{missing} part on the test set. We threshold these false detections in terms of their IoU scores  to evaluate if the missing parts are still approximately localized. We define the recall score which is the ratio between the number of detected missing part at a given IoU threshold and the total number of missing parts.
We show recall for varying IoU threshold for each method in~\fig{absent_recall}. 
For a reasonable IoU of $0.5$, RetinaNet and YOLOv3 detect approximately $20\%$ of missing parts and Faster RCNN $14\%$. Without looking at position, (IoU=0), RetinaNet and YOLOv3 detect as much as almost $80\%$ of \emph{missing} parts. Interestingly, Faster RCNN, with similar mAP object detection score as RetinaNet, detects only $32\%$ of missing parts. 
For Faster RCNN, the most hallucinated part with  $14\%$ is \emph{gear case}. For YOLOv3, a missing \emph{dynamo} is most detected and RetinaNet hallucinates most about the \emph{dress guard}.

\textbf{Evaluating visual verification.}
For visual verification, we want high recall of \emph{present} parts and low recall of \emph{missing} parts where  detecting the same object multiple times does not matter. Besides, wrongly detected \emph{missing} parts (false positives) cost more than not detected \emph{present} parts (false negatives). Thus, our $F_{vv}$ evaluation score is based on recall and inspired by the $F_{\beta}$ score~\cite{10.3115/1072064.1072067} so we can weight detection  mistakes differently as
\begin{equation}
F_{vv}=\frac{(1+\beta^2) R^P (1-R^M)}{\beta^2 (1-R^M)+R^P}.
\label{eq:fvv}
\end{equation} 
$R^P$ is the \emph{present} recall and  $R^M$ the \emph{missing} recall calculated at a certain IoU threshold. The $\beta$ parameter allows to  weight the detection mistakes, where we set the $\beta$ parameter to $0.1$ so that detections of \emph{missing} parts are 10x more costly than not detected \emph{present} parts.


\textbf{Visual verification results.} Visual verification performance is estimated by using the recall of present and missing parts.
We have two setups for visual verification calculation: with and without localization. 
\emph{Visual verification with localization:}  the \emph{present} recall has an IoU threshold of $0.5$, where the \emph{missing} recall is less relying on position and we set its IoU threshold to $0.1$. \emph{Visual verification without localization:} we set all IoU thresholds to $0$. This, in addition, allows us to evaluate a full-image  multi-class multi-label classification (MCML) approach. An Imagenet pretrained ResNet-50 architecture is fine-tuned with BCE with logits loss and SGD with an initial learning rate of $0.05$ for $15$ epochs. After every $5$ epoch, the learning rate is reduced by a factor of $10$. The network obtains $91\%$ of recall for present parts and $32\%$ of recall for missing parts.

Results are shown in~\tab{toy_results}. 
For the \emph{with localization} results, Faster RCNN outperforms RetinaNet and YOLO in terms of lower recall of \emph{missing} parts by $28\%$ and a higher $F_{vv}$ score by $72\%$.
RetinaNet and YOLOv3 detects more than $60\%$ of \emph{missing} parts and achieve only $38\%$ and $36\%$ of $F_{vv}$ score respectively.
In~\fig{bike_results4}, the AP scores of Faster RCNN and RetinaNet are quite similar, yet the $F_{vv}$ performance of Faster RCNN is almost 2 times higher than RetinaNet. RetinaNet has $7\%$ more intact recall score than YOLOv3, however, the difference for $F_{vv}$ is only $2\%$. 
For the \emph{without localization} results, when the \emph{present} and \emph{missing} IoU thresholds are set to $0$, all the methods obtain more than $90\%$ \emph{present} recall.
Interestingly, the MCML method, which only needs full image class labels, outperforms RetinaNet and YOLOv3 detectors and performs similar to Faster RCNN.

\begin{table}
    \centering
	\renewcommand{\arraystretch}{0.99}
\begin{tabular}{llllll}
\toprule
\toprule
Method      & $T^{P}$ & $T^{M}$ &$R^P$ & $R^M$ & $F_{vv}$ \\
\midrule
With localization \\
\midrule
Faster RCNN & $0.5$        & $0.1$        & $0.83$              & $\textbf{0.28}$     & $\textbf{0.72}$     \\
RetinaNet   & $0.5$        & $0.1$        & $\textbf{0.90}$     & $0.62$              & $0.38$     \\
YOLOv3      & $0.5$        & $0.1$        & $0.83$              & $0.64$              & $0.36$     \\
\midrule
\midrule
Without localization \\
\midrule
Faster RCNN & $0.0$        & $0.0$        & $0.92$              & $\textbf{0.32}$     & $\textbf{0.68}$     \\
RetinaNet   & $0.0$       & $0.0$        & $\textbf{0.99}$      & $0.79$              & $0.21$     \\
YOLOv3      & $0.0$        & $0.0$        & $0.95$              & $0.77$              & $0.23$     \\
MCML        & $0.0$        & $0.0$        & $0.91$              & $0.32$              & $0.68$    \\
\bottomrule
\end{tabular}
\caption{Visual verification of Faster RCNN, RetinaNet, YOLOv3 and MCML for different present ($T^{P}$) and missing ($T^{M}$) IoU thresholds on DelftBikes. (top) When $(T^{P}, T^{M})$ equals to $(0.5, 0.1)$: RetinaNet has highest recall for \emph{present} parts. Faster RCNN detects the fewest missing parts and has best $F_{vv}$ score. (bottom) When localization is discarded: MCML method outperforms RetinaNet and YOLOv3 and results similarly Faster RCNN in $F_{vv}$ score. 
}%
  \label{tab:toy_results}%
\end{table}

\section{Discussion and Conclusion}
We show hallucinating object detectors: Detectors can detect objects that are not in the image even with a high IoU score. We show hallucination in the context of a visual part verification task. 
We introduce DelftBikes, a novel visual verification dataset, with object class, bounding box and state labels.
We evaluate visual verification by recall, where the cost of falsely detected missing parts is more expensive than a missing present part. For object detection, Faster RCNN and RetinaNet has similar AP score, however, Faster RCNN is the better for visual verification.





One limitation of our work is that the human annotations for the non-existing parts are partly guesswork. Taking this into account, this makes it even more surprising that detectors predict with such a high IoU score.

\bibliographystyle{IEEEbib}
\bibliography{main}

\begin{thebibliography}{10}

\bibitem{rosandich1997automated}
R.~G. Rosandich,
\newblock ``Automated visual inspection systems,''
\newblock in {\em Intelligent Visual Inspection}. 1997.

\bibitem{garcia2009automated}
Hugo~C Garcia and J~Rene Villalobos,
\newblock ``Automated refinement of automated visual inspection algorithms,''
\newblock {\em IEEE T-ASE}, 2009.

\bibitem{koniar2014virtual}
Dusan Koniar, Libor Hargas, Anna Simonova, Miroslav Hrianka, and Zuzana
  Loncova,
\newblock ``Virtual instrumentation for visual inspection in mechatronic
  applications,''
\newblock {\em Procedia Engineering}, 2014.

\bibitem{kim1999visual}
Tae-Hyeon Kim, Tai-Hoon Cho, Young~Shik Moon, and Sung~Han Park,
\newblock ``Visual inspection system for the classification of solder joints,''
\newblock {\em Pattern Recognition}, 1999.

\bibitem{resendiz2013automated}
Esther Resendiz, John~M Hart, and Narendra Ahuja,
\newblock ``Automated visual inspection of railroad tracks,''
\newblock {\em IEEE transactions on ITS}, 2013.

\bibitem{marino2007real}
F.~Marino, A.~Distante, P.~L. Mazzeo, and E.~Stella,
\newblock ``A real-time visual inspection system for railway maintenance:
  automatic hexagonal-headed bolts detection,''
\newblock {\em IEEE Transactions on Systems, Man, and Cybernetics}, 2007.

\bibitem{ben2019automatic}
H.~Ben~Abdallah, I.~Jovan{\v{c}}evi{\'c}, J.-J. Orteu, and L.~Br{\`e}thes,
\newblock ``Automatic inspection of aeronautical mechanical assemblies by
  matching the 3d cad model and real 2d images,''
\newblock {\em Journal of Imaging}, 2019.

\bibitem{san2017automatic}
Marco San~Biagio, Carlos Beltran-Gonzalez, Salvatore Giunta, Alessio Del~Bue,
  and Vittorio Murino,
\newblock ``Automatic inspection of aeronautic components,''
\newblock {\em Machine Vision and Applications}, 2017.

\bibitem{baerveldt2001vision}
Albert-Jan Baerveldt,
\newblock ``A vision system for object verification and localization based on
  local features,''
\newblock {\em Robotics and Autonomous Systems}, 2001.

\bibitem{sim2006recognition}
SK~Sim, Patrick~SK Chua, ML~Tay, and Yun Gao,
\newblock ``Recognition of features of parts subjected to motion using artmap
  incorporated in a flexible vibratory bowl feeder system,''
\newblock {\em AI EDAM}, 2006.

\bibitem{ren2015faster}
Shaoqing Ren, Kaiming He, Ross Girshick, and Jian Sun,
\newblock ``Faster r-cnn: Towards real-time object detection with region
  proposal networks,''
\newblock in {\em NIPS}, 2015.

\bibitem{lin2017focal}
Tsung-Yi Lin, Priya Goyal, Ross Girshick, Kaiming He, and Piotr Doll{\'a}r,
\newblock ``Focal loss for dense object detection,''
\newblock in {\em ICCV}, 2017.

\bibitem{redmon2018yolov3}
Ali Farhadi and Joseph Redmon,
\newblock ``Yolov3: An incremental improvement,''
\newblock {\em CVPR}, 2018.

\bibitem{manfredi2020shift}
Marco Manfredi and Yu~Wang,
\newblock ``Shift equivariance in object detection,''
\newblock in {\em ECCV workshop}, 2020.

\bibitem{kayhan2020translation}
O.S. Kayhan and J.C.~van Gemert,
\newblock ``On translation invariance in {CNNs}: Convolutional layers can
  exploit absolute spatial location,''
\newblock in {\em CVPR}, 2020.

\bibitem{Barnea_2019_CVPR}
Ehud Barnea and Ohad Ben-Shahar,
\newblock ``Exploring the bounds of the utility of context for object
  detection,''
\newblock in {\em CVPR}, 2019.

\bibitem{gidaris2015object}
Spyros Gidaris and Nikos Komodakis,
\newblock ``Object detection via a multi-region and semantic segmentation-aware
  cnn model,''
\newblock in {\em ICCV}, 2015.

\bibitem{Liu_2018}
Yong Liu, Ruiping Wang, Shiguang Shan, and Xilin Chen,
\newblock ``Structure inference net: Object detection using scene-level context
  and instance-level relationships,''
\newblock {\em CVPR}, 2018.

\bibitem{zhu2017couplenet}
Yousong Zhu, Chaoyang Zhao, Jinqiao Wang, Xu~Zhao, Yi~Wu, and Hanqing Lu,
\newblock ``Couplenet: Coupling global structure with local parts for object
  detection,''
\newblock in {\em ICCV}, 2017.

\bibitem{singh2020don}
Krishna~K. Singh, Dhruv Mahajan, Kristen Grauman, Yong~Jae Lee, Matt Feiszli,
  and Deepti Ghadiyaram,
\newblock ``Don't judge an object by its context: Learning to overcome
  contextual bias,''
\newblock {\em arXiv:2001.03152}, 2020.

\bibitem{pascal-voc-2012}
M.~Everingham, L.~Van~Gool, C.~K.~I. Williams, J.~Winn, and A.~Zisserman,
\newblock ``The {PASCAL} {VOC2012},'' .

\bibitem{lin2014microsoft}
Tsung-Yi Lin, Michael Maire, Serge Belongie, James Hays, Pietro Perona, Deva
  Ramanan, Piotr Doll{\'a}r, and C~Lawrence Zitnick,
\newblock ``Microsoft coco: Common objects in context,''
\newblock in {\em ECCV}, 2014.

\bibitem{russakovsky2015imagenet}
Olga Russakovsky, Jia Deng, Hao Su, Jonathan Krause, Sanjeev Satheesh, Sean Ma,
  Zhiheng Huang, Andrej Karpathy, Aditya Khosla, Michael Bernstein, et~al.,
\newblock ``Imagenet large scale visual recognition challenge,''
\newblock {\em IJCV}, 2015.

\bibitem{DBLP:journals/corr/abs-1811-00982}
Alina Kuznetsova, Hassan Rom, Neil Alldrin, J, Jasper R.~R. Uijlings, Ivan
  Krasin, Jordi Pont{-}Tuset, Shahab Kamali, Stefan Popov, Matteo Malloci, Tom
  Duerig, and Vittorio Ferrari,
\newblock ``The open images dataset {V4:} unified image classification, object
  detection, and visual relationship detection at scale,''
\newblock {\em CoRR}, vol. abs/1811.00982, 2018.

\bibitem{chen2014detect}
Xianjie Chen, Roozbeh Mottaghi, Xiaobai Liu, Sanja Fidler, Raquel Urtasun, and
  Alan Yuille,
\newblock ``Detect what you can: Detecting and representing objects using
  holistic models and body parts,''
\newblock in {\em CVPR}, 2014.

\bibitem{stappen2020gocard}
Lukas Stappen, Xinchen Du, et~al.,
\newblock ``Go-card -- generic, optical car part recognition and detection:
  Collection, insights, and applications,'' 2020.

\bibitem{he2016deep}
Kaiming He, Xiangyu Zhang, Shaoqing Ren, and Jian Sun,
\newblock ``Deep residual learning for image recognition,''
\newblock in {\em CVPR}, 2016.

\bibitem{10.3115/1072064.1072067}
N.~Chinchor,
\newblock ``Muc-4 evaluation metrics,''
\newblock in {\em MUC4}. 1992, Association for Computational Linguistics.

\end{thebibliography}
\end{document}